# A Harmonic Potential Approach for Simultaneous Planning & Control of a Generic UAV Platform

Ahmad A. Masoud

*Electrical Engineering, King Fahad University of Petroleum & Minerals, Dhahran Saudi Arabia 31261* , e-mail: masoud@kfupm.edu.sa

***Abstract:*** Simultaneous planning and control of a large variety of unmanned aerial vehicles (UAVs) is tackled using the harmonic potential field (HPF) approach. A dense reference velocity field generated from the gradient of an HPF is used to regulate the velocity of the UAV concerned in a manner that would propel the UAV to a target point while enforcing the constraints on behavior that were *a priori* encoded in the reference field. The regulation process is carried-out using a novel and simple concept called the: virtual velocity attractor (VVA). The combined effect of the HPF gradient and the VVA is found able to yield an efficient, easy to implement, well-behaved and provably-correct context-sensitive control action that suits a wide variety of UAVs. The approach is developed and basic proofs of correctness are provided along with simulation results.

**Nomenclature:**

| | | |
|---|---|---|
| $P$ | = | a point in an abstract N-dimensional space (usually N=3, P=[x y z]$^t$), |
| $P_s$ | = | the starting point, |
| $P_r$ | = | the reference or target point, |
| $V$ | = | potential field |
| $M$ | = | the point mass of the UAV, |
| $F_T$ | = | the resultant force along the velocity vector, |
| $F_N$ | = | the resultant force normal to the velocity vector, |
| $g$ | = | the constant of gravity, |
| $T$ | = | the thrust from the UAV engine, |
| $D$ | = | the aerodynamic drag, |
| $L$ | = | the aerodynamic lift, |
| $K_u, K\lambda$ | = | are positive constants, |
| $\Omega$ | = | the workspace, |
| $\Gamma$ | = | boundary of the forbidden regions (obstacles), |
| $\Omega`$ | = | the region in which directional constraints should be enforced, |
| $\beta(P)$ | = | a differentiable function that describe in a probabilistic manner everywhere in $\Omega$ the fitness of a point P for motion to pass through, |
| $\nu$ | = | the radial speed of the UAV, |
| $\gamma$ | = | flight path angle, |
| $\psi$ | = | directional angle, |
| $\lambda$ | = | a vector describing motion in the local coordinates of the UAV, $\lambda=[\nu\ \gamma\ \psi]^t$ , |
| $\sigma$ | = | the banking angle, |
| $\epsilon$ | = | the angel of attack, |
| $G(\lambda)$ | = | An orthogonal coordinate transformation between the local coordinates of the UAV and its global coordinates, |
| $F(\lambda,u)$ | = | the local coordinates state actuation function, |
| $\eta_\lambda$ | = | the minimum eigenvalue of $Q_\lambda$ ($\eta_\lambda$>0 if the system is fully or redundantly actuated), |
| $\eta_P$ | = | the minimum eigenvalue of the matrix $Q_P$, |
| $\Lambda$ | = | a vector encoding the directional constraints, |
| $\sigma_d(P)$ | = | directionally sensitive fitness measure, |
| $\sigma_f$ | = | value of $\sigma_d$ when motion is in accordance with $\Lambda$, |
| $\sigma_b$ | = | value of $\sigma_d$ when motion violates $\Lambda$, |
| $C_L, C_D$ | = | positive constants, |
| $\rho$ | = | air density. |
| $\chi$ | = | barrier control signal |
| $\Xi, \dot{\Xi}$ | = | Liapunov function and its derivative, unconstrained system |
| $\Xi_C, \dot{\Xi}_C$ | = | Liapunov function and its derivative, constrained system |
| $\Omega_u$ | = | feasible subset of the control space |
| $\Gamma_u$ | = | boundary of $\Omega_u$ |
| $\Omega_s$ | = | the set in $P \times \lambda \times u$ where $u \in \Omega_u$ |
| $\Gamma_s$ | = | boundary of $\Omega_s$ |

## I. Introduction

The past decade has witnessed a surge in demand for unmanned aerial vehicles (UAVs) to perform critical tasks such as: search and rescue, reconnaissance and target tracking [1]-[3]. Although the hardware for these robotics agents is becoming commercially available in many different forms (figure-1a) at reasonable prices, the software needed to allow reliable, de-skilled operation of these agents is still the focus of intensive study and development [4]-[6]. It is not uncommon to see form-specific controllers capable of working with only one design while failing to work with others. It is highly unlikely that a controller designed for a fixed-wing UAV [7]-[9] to work with a helicopter-type [10]-[12] or a controller designed to work for a helicopter UAV to properly function with a quad-rotor UAV [13]-[15] tilt rotor [16]-[18] or other types of UAVs [19]-[21]. This is understandable, since all these agents are severely nonlinear dynamic systems that are subject to nonholonomic constraints making controller design a challenging task.

For these agents to perform a task, a specific type of intelligent, goal/mission-oriented controllers that have the ability to embed the UAV in a given context is needed. These controllers are usually referred to as: navigation controllers (NC) or kinodynamic motion planners. Managing the hierarchies of functions needed to support a UAV is being approached by researchers at different levels of the problem with a focus that is limited to individual components of the system or a wider one that aims at integrating more than one essential component in a functioning subsystem of the UAV. Classical controllers that allow a user to direct the UAV along a desired orientation and radial speed were suggested in [22],[23]. A generalization that would allow a UAV to track a target or a reference trajectory was suggested in [24],[25]. Another approach to tackle the problem is to focus only on the kinematic aspects using a planner to translate the context, goal, and mission constraints into a spatial trajectory [26],[27], the existence of a controller that is capable of tracking the trajectory is assumed. The difficult task of joint design of planning and control (navigation control; NC) was attempted with varying degrees of success in [28]-[30]. Work on the design of modular structures that aim at full system integration may be found in [31],[32].

Despite the intensive effort to develop such controllers and the significant advances achieved there is still a long list of requirements that needs to be addressed. For example almost all of the available controllers are involved, are not easy to tune and require too much processing power. It is desired that the controllers be simple, yet robust, and easy to operate and tune. The controller should also be able to impose a diverse set of constraints in both the workspace of the UAV and in its control space. The ability to integrate, in a provably-correct manner, planning and control is highly desirable if not a must. Due to the high cost of the UAV control software, it is also desirable that the controller accommodate, with minor adjustments, a variety of UAVs.

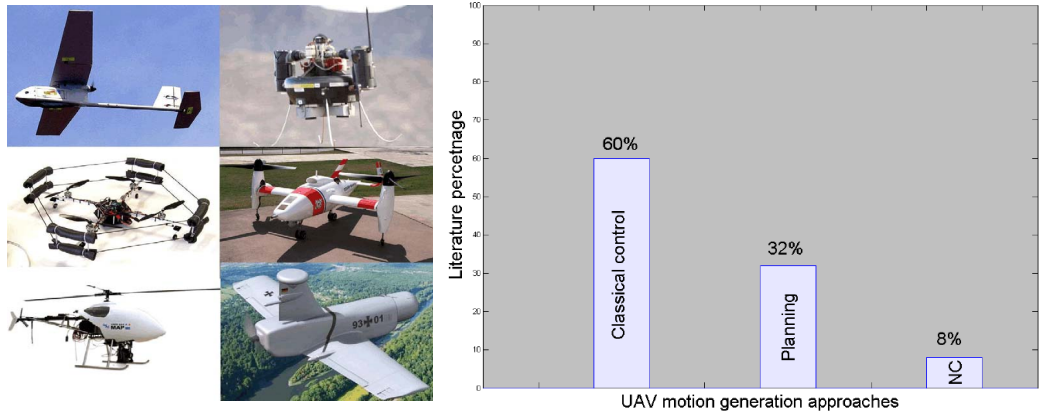

Figure-1: a: Different types of UAVs, b: Estimated literature of different approaches for UAV motion generation.

This paper attempts to jointly address some of the above requirements in a practical manner. Its an extension to the initial results reported in [50]. It develops a flexible, easy to tune, generic navigation controller that is applicable to a wide range of UAVs. Provably-correct navigation controllers for UAVs are found to be an effective way for the generation of intelligent, context-sensitive, goal-oriented behavior. It is a relatively recent and challenging area that is still in need of a considerable amount of development (figure-1b). This type of context-sensitive control is known to have several advantages over the widely used high level-low level context-sensitive control scheme in which a planning stage (high level controller) is connected to a classical controller (low level control). The suggested approach combines an effective and versatile motion planning technique called the harmonic potential field (HPF) motion planner [33]-[34], the attractor potential field approach originally suggested by Khatib [35] along with a two-stage model for UAVs. The guidance field from the HPF planner is used to provide the reference velocity field which the UAV must enforce if it is to execute the mission in the desired manner. The attractor field approach along with the two stage model in [43] are combined to work as a virtual velocity attractor (VVA) that would attempt, at an arbitrary point in space, to make the velocity of the UAV coincide with the reference velocity.

## II. The HPF Approach: A Background

The harmonic potential field approach is a powerful, versatile and provably-correct means of guiding motion in an N-dimensional abstract space to a goal state subject to a set of constraints that is used to represent an environment. The approach works by converting the goal, representation of the environment and constraints on behavior into a reference velocity vector field (figure-2). This reference field is usually generated from a properly conditioned negative gradient of an underlying potential field. A basic setting of the HPF approach is shown below (1):

solve: $\nabla^2 V(P) \equiv 0 \quad P \in \Omega$ (1)

subject to: $V(P) = 1$ at $P = \Gamma$ and $V(P_r) = 0$,

A provably-correct path may be generated using the gradient dynamical system:

$$\dot{P} = -\nabla V(P). \quad (2)$$

Many variants of the above setting were later proposed to extend the capabilities of the HPF approach. For example, it is demonstrated that the approach can be used for planning in complex unknown environment [36] relying on local sensing only (figure-3),

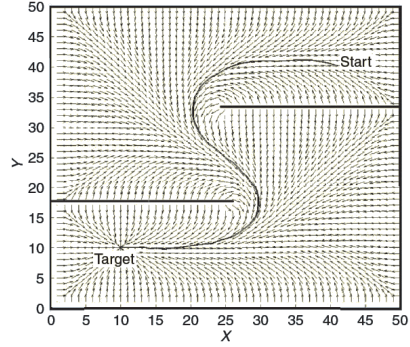

Figure-2: The reference velocity field from an HPF.

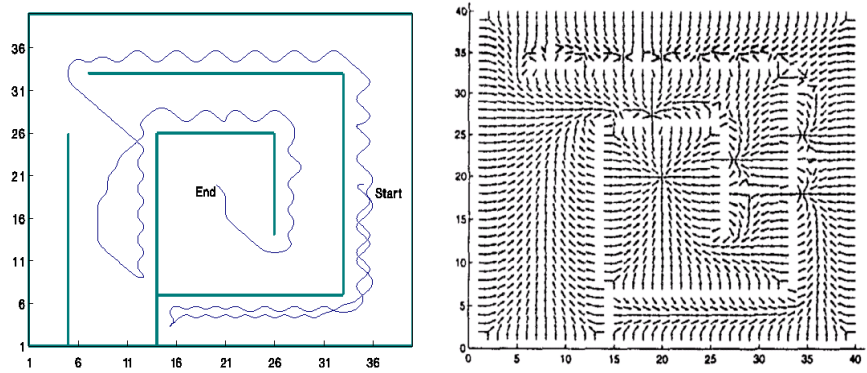

Figure-3: planning in unknown environments

The HPF approach can incorporate directional constraints along with regional avoidance constraints [37] in a provably-correct manner to plan a path to a target point (figure-4). The navigation potential may be generated using the boundary value problem (BVP) :

jointly solve: $\quad\quad\quad \nabla^2 V(P) \equiv 0 \quad\quad P \in \Omega - \Omega' \quad\quad (3)$
and $\quad\quad\quad\quad\quad \nabla \cdot [\Sigma(P) V(P)] = 0 \quad\quad P \in \Omega'$
subject to: $\quad\quad V(P) = 1$ at $P = \Gamma$ and $V(P_r) = 0$

where

$$\Sigma(P) = \begin{bmatrix} \sigma(p) & 0 & \ldots & 0 \\ 0 & \sigma(p) & \ldots & 0 \\ & & \cdot & \\ 0 & 0 & \ldots & \sigma(p) \end{bmatrix}$$

$$\sigma_d(P) = \begin{bmatrix} \sigma_f & -\nabla V(P)^t \Lambda(P) > 0 \\ \sigma_b & -\nabla V(P)^t \Lambda(P) \leq 0 \end{bmatrix}$$

A provably correct trajectory to the target that enforces both the regional avoidance and directional constraints may be simply obtained using the gradient dynamical system in (2).

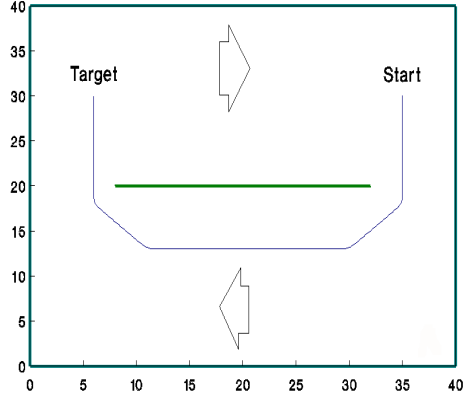
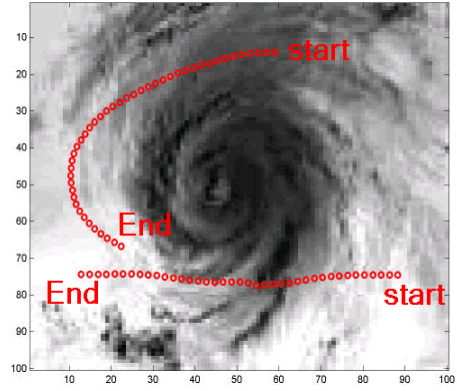

Figure-4: Directional & regional avoidance constraints

Figure-5: Planning in non-divisible environments

The HPF approach may be modified to take into consideration inherent ambiguity [38] that prevents the partitioning of an environment into admissible and forbidden regions (figure-5). The BVP that generates the navigation potential for this case is:

solve $\quad\quad\quad\quad\quad\quad\quad \nabla \cdot (\beta(P)\nabla V(P)) \equiv 0 \quad\quad P \in \Omega \quad\quad\quad (4)$
subject to: $\quad\quad\quad\quad\quad V(P_s) = 1, \ V(P_r) = 0$

A provably correct path that avoids definite threat regions ($\beta(P)=0$) and converge to the target may be generated using the gradient dynamical system in (2). The HPF-based approach in [38] may be easily modified to take advantage of a drift field that my be present in an environment [39] in a manner that suits planning for energy exhaustive missions (figure-6). It was also demonstrated in [40] that the HPF approach can work with integrated navigation systems that can efficiently function in a real-life situation (figure-7). Work on extending the HPF approach to work with dynamical and nonholonomic systems may be found in [41]-[43]. An HPF-based, decentralized, Multi-agent approach was suggested in [47].

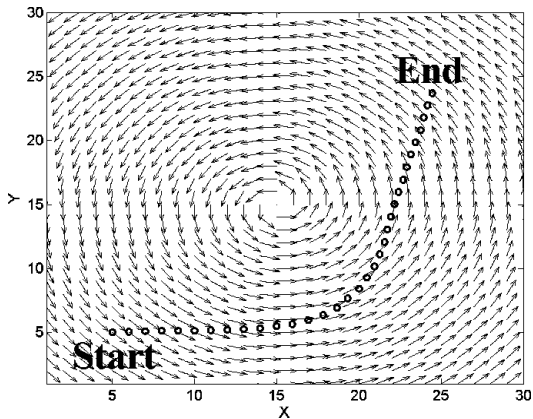
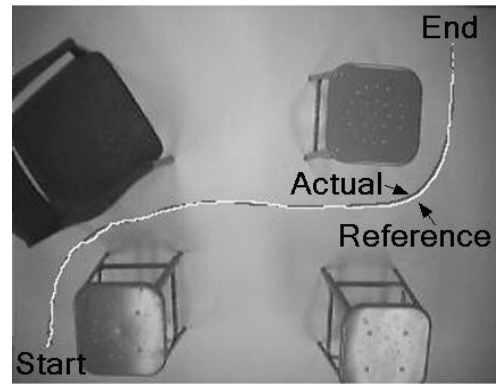

Figure-6: planning in the presence of drift fields

Figure-7: Integrated, HPF-based navigation

### III. A two-stage model

A two-stage model to describe motion of a mobile robot was suggested in [43]. The model is based on dividing a robot into a local actuation stage that couples the control signal to the variables describing the robot's motion in its local coordinates and a global stage that transforms the local variables into world-coordinate motion descriptors. The model, coupled with the HPF approach, was shown to be an effective means for planning motion for mobile robots in both the kinematic and kino-dynamic cases. However the work in [43] is based on the assumption that the local motion actuation stage is invertible. In this work it is shown that the above combination can be effectively utilized in the case where the relation between the control (u) variables and the local motion descriptors ($\lambda$) is non-invertible.

A model that suits most (if not all) UAVs have the form:

$$\begin{aligned} \dot{P} &= G(\lambda) \\ \dot{\lambda} &= F(\lambda, u) \end{aligned} \tag{5}$$

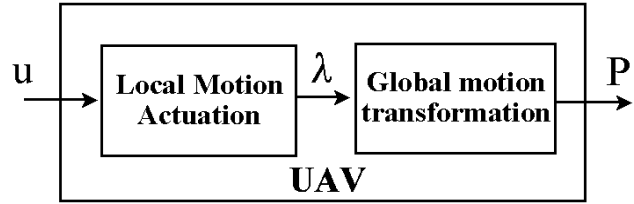

Figure-8: A two-stage model for UAVs

It ought to be noticed that the system equations above do apply to other types of robots such as autonomous under water vehicles (AUVs) [44] and spherical robots [45]. A specific form for equation (5) that describe a fixed-wing (figure-9) aircraft [46] is:

$$\begin{aligned} \dot{x} &= v \cdot \cos(\gamma)\cos(\psi) \\ \dot{y} &= v \cdot \cos(\gamma)\sin(\psi) \\ \dot{z} &= v \cdot \sin(\gamma) \\ \dot{v} &= \frac{F_T}{M} - g \cdot \sin(\gamma) \\ \dot{\gamma} &= \frac{F_N \cdot \cos(\sigma)}{M \cdot v} - g\frac{\cos(\gamma)}{v} \\ \dot{\psi} &= \frac{F_N \cdot \sin(\sigma)}{M \cdot v \cdot \cos(\gamma)}. \end{aligned} \tag{6}$$

$$F_T = T \cdot \cos(\epsilon) - D, \qquad F_N = T \cdot \sin(\epsilon) + L \tag{7}$$

$$D = \frac{C_D}{2}\rho v^2, \qquad L = \frac{C_L}{2}\rho v^2 \tag{8}$$

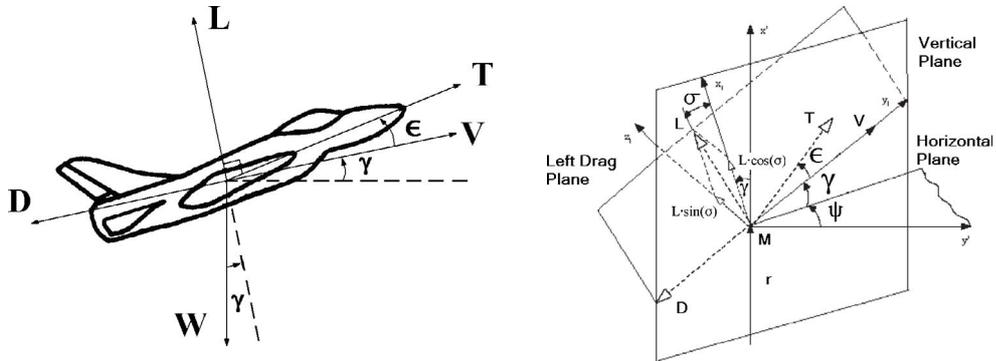

Figure-9: A fixed-wing UAV.

### IV. The HPF-VVA Approach

An HPF-based technique guides motion to a target point and orientation in a provably-correct manner that observes a set of *a priori* specified set of constraints by converting the mission data into a dense vector field that covers the workspace of the agent. This reference field provides at each point the reference velocity instruction that the robot needs to abide by in order for the mission to be accomplished. This process is provably-correct for a massless, single integrator, holonomic system. While it may appear that the capabilities of the HPF approach falls way below the minimum needed to handle the dynamic system in (5), the approach has properties that are adaptable for use with severely nonlinear systems such as UAVs. The reference velocity field generated by an HPF method is a region to point planner. In other words successfully executing any guidance instruction irrespective of its location in space will drive the UAV closer to its goal while upholding the constrains. Moreover, the solution trajectories the HPF approach generates are analytic and expected to be well-behaved when dynamics and nonholonomicity are considered. Therefore if at a point **P** in space the velocity of the UAV ($\dot{\mathbf{P}}$) is driven to coincide with the velocity reference from the HPF planner ($\dot{\mathbf{P}}_r = -\nabla V(\mathbf{P})$), the actual trajectory of the UAV will converge immediately or after a short transient period to the provably-correct trajectory generated by an HPF planner. This may be implemented by constructing an artificial force $\mathbf{F_P}$ that attempts to attract the velocity of the UAV to the desired velocity from the HPF planner (figure-10)

$$\mathbf{F_P} = K_\lambda \cdot (\dot{\mathbf{P}}_r - \dot{\mathbf{P}}) = K_P \cdot \dot{\mathbf{P}}_e . \qquad (9)$$

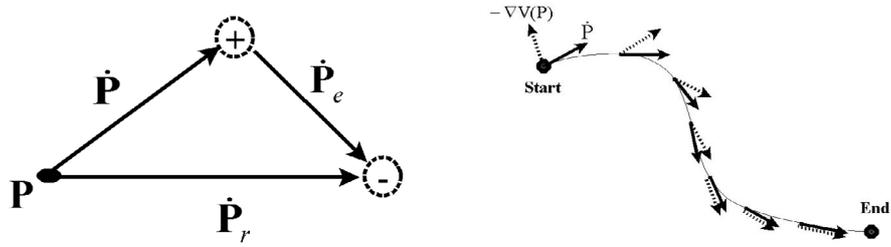

Figure-10: Linear velocity attracor

Since the local motion vector $\lambda$ of the UAV is what causes its velocity in the world coordinate to change ($\dot{\mathbf{P}} = \mathbf{G}(\lambda)$), a force $\mathbf{F}_\lambda$ in the $\lambda$ coordinates whose effect is equivalent to $\mathbf{F_P}$ has to be constructed using force transformation (10):

$$\mathbf{F}_\lambda = \mathbf{J}_\lambda^T \mathbf{F_P}$$

where
$$\mathbf{J}_\lambda = \frac{\partial \mathbf{G}(\lambda)}{\partial \lambda} \qquad (10)$$

The fictitious force $\mathbf{F}_\lambda$ may be used as the desired velocity ($\dot{\lambda}_r = \mathbf{F}_\lambda$) in the UAV's local coordinates ($\lambda$). In a manner similar to the above, another artificial force is constructed so that at each point in the coordinates ($\lambda$) the local velocity of the UAV ($\dot{\lambda}$) is driven to coincide with the reference velocity $\dot{\lambda}_r$ (figure-11). This artificial force ($F_u$) may be chosen as the scaled error between the two local velocities:

$$\dot{\lambda}_e = \dot{\lambda}_r - \dot{\lambda}$$
$$F_u = k_u \cdot \dot{\lambda}_e . \qquad (11)$$

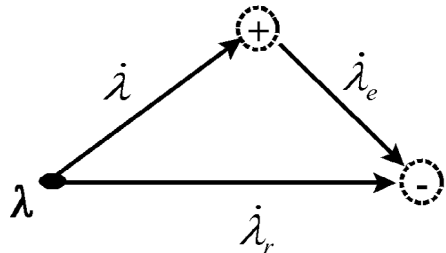

Figure-11: Local velocity attractor.

The artificial force in the $\lambda$ coordinates must be transformed to its equivalent in the control variable coordinates (u). The control coordinate force is used to direct the change of the control signal:

$$\dot{u} = K_u \cdot J_u^T \cdot \dot{\lambda}_e$$

where
$$J_u = \frac{\partial F(\lambda, u)}{\partial u} \qquad (12)$$

the control signal fed to the actuators of the UAV may be simply derived as:

$$u(t) = \int_{t_0}^{t} \dot{u}\, dt \qquad (13)$$

The block diagram of the overall control structure is shown in figure-12. The control signal of the UAV is generated by jointly solving the nonlinear dynamical systems in (14,5):

$$\begin{aligned}
\dot{u} &= K_u J_u^T \left[\dot{\lambda}_r - \dot{\lambda}\right] \\
&= K_u J_u^T \left[K_\lambda J_\lambda^T (-\nabla V(P) - \dot{P}) - \dot{\lambda}\right] \\
&= K_u J_u^T \left[K_\lambda J_\lambda^T (-\nabla V(P) - G(\lambda)) - F(\lambda, u)\right] \\
&= Q(P, \lambda, u)
\end{aligned} \qquad (14)$$

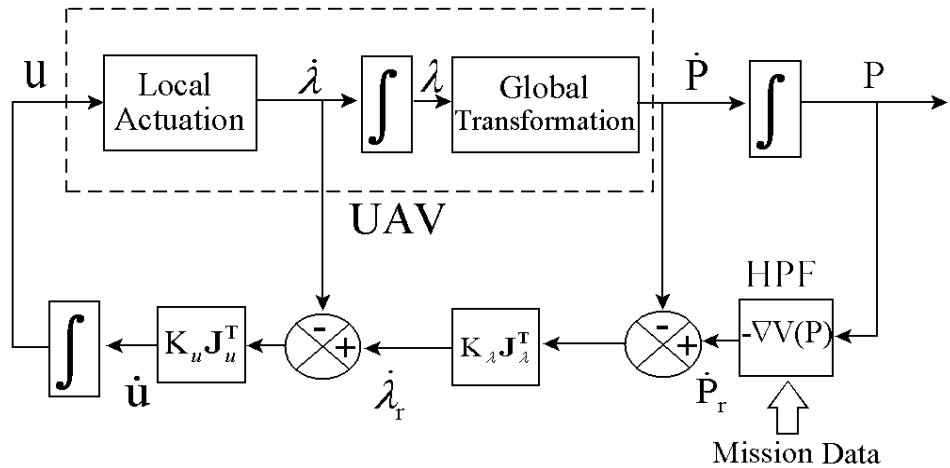

Figure-12: Suggested navigation control.

As can be seen the VVA approach treats control space in a manner that is similar to workspace where the control signal is viewed as a point moving in space under the influence of a force. This makes it possible to easily apply constraints on the control signal hence jointly constraining state and control spaces. This is achieved by simply augmenting the control signal with a barrier control ($\chi$) to keep u in the feasible part of the control space ($\Omega_u$):

$$\dot{u} = Q(P, \lambda, u) + \chi(u) \qquad (15)$$

The simplest and most practical geometry $\Omega_u$ may assume is a convex rectangular region of the form:

$$\Omega_u = \{u : u_i^- \leq u_i \leq u_i^+, i = 1,..M\} \qquad (16)$$

where $u_i^+$ and $u_i^-$ are the upper and lower bounds on $u_i$ respectively. The barrier control signal is localized to the boundary of $\Omega_u$ ($\Gamma_u = \partial \Omega_u$). The barrier control used with the i'th component of u ($\chi_i$) may be constructed as:

$$\chi_i(u) = \begin{bmatrix} -K & u_i = u_i^+ \\ +K & u_i = u_i^- \\ 0 & \text{elsewhere} \end{bmatrix}, \quad i=1,..,M \qquad (17)$$

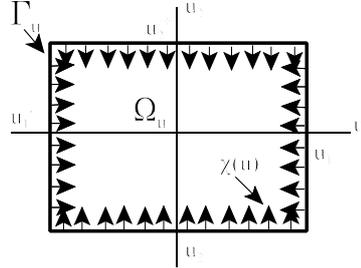

Figure-13: Barrier control in a 2D control space.

where K is a positive constant. The full barrier control may be expressed as the algebraic sum of the individual barrier controls:

$$\chi(u) = \sum_{i=1}^{M} \chi_i(u_i). \qquad (18)$$

In the following section a proof is provided of the ability of the control signal to guarantee stability of the UAV. It is shown that the controller can force the dynamical trajectory of the UAV to mimic the kinematic trajectory from the HPF planner; hence guaranteeing convergence to the target while upholding the constraints encoded in the HPF-generated kinematic trajectory. It is also shown that the dynamic trajectory converges instantly to the kinematic trajectory or, at worst, has an exponential convergence rate which may be directly controlled by $K_u$ and $K_\lambda$. The addition of constraints in the control space is proven not to affect convergence where under simple and nonastringent conditions the constrained system is guaranteed to converge to the target point and configuration while maintaining the control signal confined to $\Omega_u$ at all time.

## V. Correctness analysis

This section examines the behavior of the suggested controller.

**proposition-1:** The controller suggested in (14) is unconditionally stable so that for any $K_u>$ and $K_\lambda>0$:

$$\lim_{t \to \infty} \dot{P}_e \to 0. \qquad (19)$$

**Proof:** First, the controller is considered to be fast acting so that at a fixed point in space (P) the velocity of the UAV is attracted to a static reference supplied by the negative gradient of the HPF. Proof of the above proposition may be established at two stages: first the error function (20) in the local coordinates of the UAV is constructed:

$$E_{\dot\lambda}(t) = \|\dot\lambda_e\|^2 = \|\dot\lambda_r - \dot\lambda\|^2. \qquad (20)$$

Keeping in mind that at a specific point in space, the reference may be considered static, we have:

$$\frac{d}{dt}E_{\dot\lambda}(t) = -2\dot\lambda_e^T \ddot\lambda. \qquad (21)$$

From (5) we have:
$$\ddot\lambda = \frac{\partial F}{\partial u}\dot u \qquad (22)$$

$$\frac{d}{dt}E_{\dot\lambda}(t) = -2\dot\lambda_e^T \frac{\partial F}{\partial u}\dot u \qquad (23)$$

substituting equation (14) in (23), we have:

$$\frac{d}{dt} E_\lambda(t) = -2K_u \dot{\lambda}_e^T (\frac{\partial F}{\partial u} \frac{\partial F}{\partial u}^T) \dot{\lambda}_e = -2K_u \dot{\lambda}_e^T Q_\lambda \dot{\lambda}_e \qquad (24)$$

The matrix:
$$\frac{\partial F}{\partial u} \frac{\partial F}{\partial u}^T \geq 0 \qquad (25)$$

is at least positive semi-definite. If the rank of the Jacobian matrix is equal to its rows (i.e. the system is fully or redundantly actuated), the matrix in (25) is positive definite which for this case implies that:

$$\frac{d}{dt} E_\lambda(t) < 0$$

and leads to:
$$\lim_{t \to \infty} \dot{\lambda}_e \to 0. \qquad (26)$$

In a similar way as above let the following error function be constructed:
$$E_{\dot{P}}(t) = \|\dot{P}_e\|^2 = \|\dot{P}_r - \dot{P}\|^2. \qquad (27)$$

$$\frac{d}{dt} E_{\dot{P}}(t) = -2\dot{P}_e \ddot{P}.$$

Since:
$$\ddot{P} = \frac{\partial G}{\partial \lambda} \dot{\lambda} \qquad (28)$$

and
$$\lim_{t \to \infty} \dot{\lambda} \to \dot{\lambda}_r \qquad (29)$$

we have:
$$\ddot{P} = \frac{\partial G}{\partial \lambda} \dot{\lambda}_r = K_P \frac{\partial G}{\partial \lambda}^T \frac{\partial G}{\partial \lambda} \dot{P}_e \qquad (30)$$

Therefore:
$$\frac{d}{dt} E_{\dot{P}}(t) = -2K_\lambda \dot{P}_e^T (\frac{\partial G}{\partial \lambda}^T \frac{\partial G}{\partial \lambda}) \dot{P}_e = -2K_\lambda \dot{P}_e^T Q_P \dot{P}_e \qquad (31)$$

Since G is an orthogonal coordinate transformation, $(\partial G/\partial \lambda)$ is full rank and the matrix:

$$\frac{\partial G}{\partial \lambda}^T \frac{\partial G}{\partial \lambda} > 0 \qquad (32)$$

is positive definite, i.e.
$$\frac{d}{dt} E_{\dot{P}}(t) < 0 \qquad (33)$$

is negative definite or equivalently:
$$\lim_{t \to \infty} \dot{P}_e \to 0. \qquad (34)$$

**proposition-2:** If $\dot{P}_e(0) = \dot{\lambda}_e(0) = 0$, then $\dot{P}_e(t) = \dot{\lambda}_e(t) = 0 \, \forall \, t$. (35)

**proof:** The $\dot{\lambda}_e$ error measure rate of change may be bounded as:

$$\frac{d}{dt} E_\lambda(t) = -2K_u \dot{\lambda}_e^T Q_\lambda \dot{\lambda}_e \leq -2K_u \eta_\lambda \dot{\lambda}_e^T \dot{\lambda}_e = -2K_u \eta_\lambda E_\lambda(t). \qquad (36)$$

Therefore an upper bound on $E_\lambda(t)$ may be constructed as:
$$E_\lambda(t) \leq \exp(-2K_u \eta_\lambda t) E_\lambda(0) \qquad (37)$$

As can be seen if $\dot{\lambda}_e(0) = 0$ then $E_\lambda(0) = 0$. This in turns imply that $E_\lambda(t) = 0 \, \forall \, t$ which leads to $\dot{\lambda}_e(t) = 0 \, \forall \, t$. This enables us to bound the error measure rate of change on $\dot{P}_e$ as:

$$\frac{d}{dt} E_{\dot{P}}(t) = -2K_\lambda \dot{P}_e^T Q_P \dot{P}_e \leq -2K_\lambda \eta_P \dot{P}_e^T \dot{P}_e = -2K_\lambda \eta_P E_{\dot{P}}(t). \qquad (38)$$

As before, the error measure on $\dot{P}_e$ may be bounded as:
$$E_{\dot{P}}(t) \leq \exp(-2K_\lambda \eta_P t)E_{\dot{P}}(0) \tag{39}$$
As can be seen if $\dot{P}_e(0)=0$ then $E_{\dot{P}}(0) = 0$. This implies that $E_{\dot{P}}(t) = 0 \ \forall \, t$ which leads to $\dot{P}_e(t)=0 \ \forall \, t$. Even if the initial values of the errors are not equal to zero, the error measures will still exponentially decay with time leading to fast alignment of the velocity vector of the UAV with the reference velocity vector from the gradient of the harmonic potential.

**Proposition-3:** let $\rho_k$ be the provably-correct, convergent and constraints-satisfying kinematic trajectory obtained from the gradient of the harmonic potential planner:
$$\rho_k = \{P(t): \dot{P} = -\nabla V(P)\}. \tag{40}$$
Let $\rho_d$ be the dynamic trajectory of the UAV:
$$\rho_d = \left\{P(t): \begin{bmatrix} \dot{P} \\ \dot{\lambda} \\ \dot{u} \end{bmatrix} = \begin{bmatrix} G(\lambda) \\ F(\lambda, u) \\ Q(P, \lambda, u) \end{bmatrix}\right\} \tag{41}$$

if $\dot{P}_e(0) = \dot{\lambda}_e(0) = 0$, then $\rho_d = \rho_k \ \forall t$.

**Proof:** this result directly follows from proposition-2.

**Proposition-4:** Let
$$u_r = \lim_{t \to \infty} u(t) \tag{42}$$

resulting from the system :
$$\begin{bmatrix} \dot{P} \\ \dot{\lambda} \\ \dot{u} \end{bmatrix} = \begin{bmatrix} G(\lambda) \\ F(\lambda, u) \\ Q(P, \lambda, u) \end{bmatrix} \tag{43}$$

If
$$u_r \in \Omega_u \tag{44}$$
$$K \geq \underset{P, \lambda, u}{\text{Max}} |Q(P, \lambda, u)| \tag{45}$$

and $u_r$ is unique at least in $\Omega_u$, then for the system
$$\begin{bmatrix} \dot{P} \\ \dot{\lambda} \\ \dot{u} \end{bmatrix} = \begin{bmatrix} G(\lambda) \\ F(\lambda, u) \\ Q(P, \lambda, u) + \chi(u) \end{bmatrix} \tag{46}$$

we have:
$$\begin{array}{l} \lim_{t \to \infty} P(t) \to P_r \\ \lim_{t \to \infty} \lambda(t) \to \lambda_r \\ \lim_{t \to \infty} u(t) \to u_r \end{array} \tag{47}$$

and $u(t) \in \Omega_u \ \forall t$.

**Proof:** Let $R \in P \times \lambda \times u$ and the point $R_T = [P_r \ \lambda_r \ u_r]^t$. Since for the unconstrained system we have:
$$\lim_{t \to \infty} R(t) \to R_T \tag{48}$$
there exist a Liapunov function $\Xi(R)$ such that:
$$\begin{array}{ll} \Xi(R) > 0 & \forall R \\ \Xi(R) = 0 & R = R_T \end{array} \tag{49}$$
such that:
$$\begin{array}{ll} \dot{\Xi}(R) < 0 & \forall R \\ \dot{\Xi}(R) = 0 & R = R_T \end{array}$$

Let $\Omega_s$ be an open subset in $R(\Omega_s \subset R)$ such that:
$$\Omega_S = \{R: u \in \Omega_u\} \tag{50}$$
and $\Gamma_s$ be its boundary ($\Gamma_s = \partial\Omega_s$).

It is not hard to show that if K is selected using equation-45, the barrier function will be able to keep u in $\Omega_u$ for all t. In other words, R will always be in $\Omega_s \cup \Gamma_s$ for all t. Let $\Xi_c(R)$ be a Liapunov function for the constrained system in (46), where $\Xi_c(R) = \Xi(R)$ $\forall \in \Omega_s \cup \Gamma_s$. If $R \in \Omega_s$, i.e. $\chi(u)=0$, we have:

$$\dot{\Xi}_C(R) = \dot{\Xi}(R). \tag{51}$$

On the other hand, if $R \in \Gamma_s$, then one or more components of the control vector u is kept to a constant value by the barrier control $\chi$. This means that the derivative of $F(\lambda, u)$ with respect to these components is equal to zero. This results in some of the rows of $J_u$ being zeros and could lead to the matrix $J_u^T J_u$ becoming negative semi-definite for the constrained system:

$$\dot{\Xi}_C(R) \leq 0 \qquad R \in \Omega_s \cup \Gamma_s. \tag{52}$$

According to the LaSalle invariance principle [49], R(t) will converge to the minimum invariant set. This set may be obtained as the solution of equation-53:

$$\begin{bmatrix} G(\lambda) \\ F(\lambda, u) \\ Q(P, \lambda, u) + \chi(u) \end{bmatrix} = 0. \tag{53}$$

By noting that the vector $G(\lambda)$ cannot equal to zero unless the radial velocity $v$ is equal zero, one concludes that it is impossible for the set $\Gamma_s$ to be a part of the minimum invariant set. This will only leave $R_T$ as the only point in this set. Therefore:

$$\lim_{t \to \infty} R(t) \to R_T \tag{54}$$

## VI. Simulation Results

In this section the capabilities of the suggested navigation control scheme are demonstrated by simulation.

### *Fixed wing UAV:*

The navigation control is tested for the UAV model in (4). For this case we have:

$$J_\lambda = \begin{bmatrix} C\gamma C\psi & -v \cdot S\gamma C\psi & -v \cdot C\gamma S\psi \\ C\gamma S\psi & -v \cdot S\gamma C\psi & v \cdot C\gamma C\psi \\ S\gamma & v \cdot C\gamma & 0 \end{bmatrix} \tag{55}$$

$$J_u = \begin{bmatrix} \dfrac{1}{M} & 0 & 0 \\ 0 & \dfrac{C\sigma}{M \cdot v} & \dfrac{-F_N}{M \cdot v} \cdot S\sigma \\ 0 & \dfrac{S\sigma}{M \cdot vC\gamma} & \dfrac{F_N}{M \cdot v} \cdot \dfrac{C\sigma}{C\gamma} \end{bmatrix}$$

where $C(*) = \cos(*)$, $S(*) = \sin(*)$, $M=1$, $K_u=1$, $K_\lambda=2$.

In the first case the controller is required to fly the UAV at a constant speed ($v_r=1$) along the x direction maintaining y=z=2 starting from the initial position x=y=z=0 and initial configuration $v=0$, $\gamma=0$ $\psi=\pi/4$. Figure-14 shows the spatial trajectory generated by the navigation control and figure -15 shows the corresponding x,y,z of the trajectory versus time. As can be seen, the trajectory is smooth and well-behaved. The radial velocity of the UAV (figure-16) quickly settles in a well-behaved manner to the desired radial speed. The orientation angles of the UAV ($\gamma,\psi$) as a function of time have a smooth well-behaved profile (figure-17). The control variables: banking angle ($\sigma$), normal force ($F_N$) and resultant force along $v$ ($F_T$) are shown in figures-18,19,20 respectively. As can be seen the control signals are bounded and well-behaved.

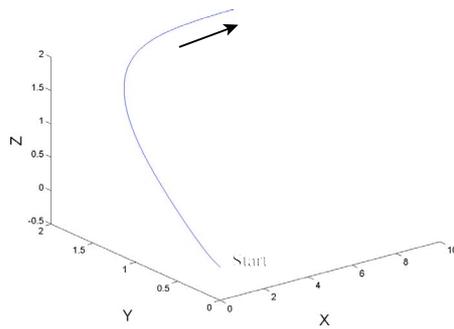

Figure-14: Spatial trajectory UAV

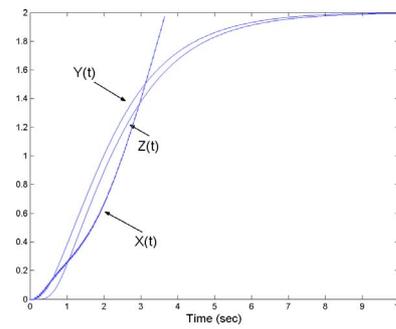

Figure-15: The xyz components of the trajectory

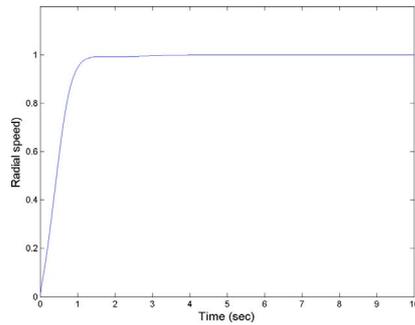

Figure-16: Radial speed of the UAV

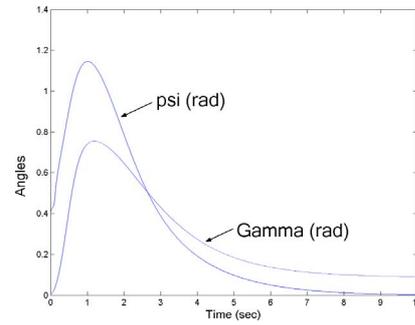

Figure-17: Orientation of the UAV

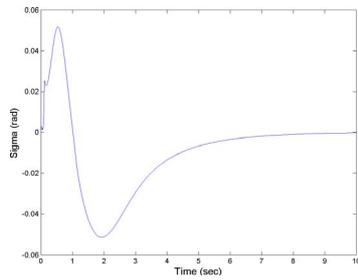

Figure-18: Banking angle

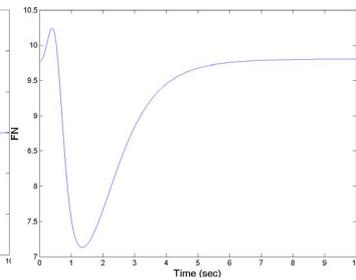

Figure-19: Normal force

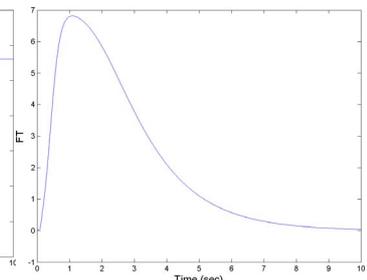

Figure-20: Tangent force

In figure-21 the control is tested for the multi-UAV case. An antipodal configuration is used to set the UAVs on a potential collision course. Both UAVs are equipped with the suggested navigation control. One UAV is non-cooperative and is treated by the other as an obstacle. As can be seen from the inter-distance curve (figure-22), collision was avoided and each UAV proceeded safely towards its destination. The radial speeds of both UAVs are shown in figures-23,24. It is worth noting that the radial velocity of the maneuvering UAV remained around the rated velocity during the evasion maneuver.

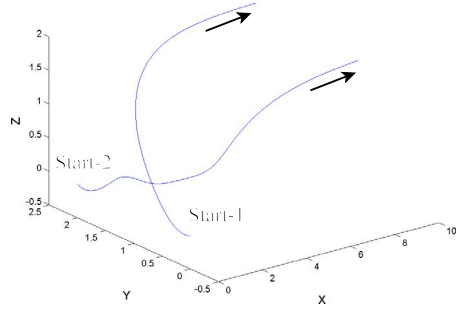

Figure-21: Trajectories of the UAVs

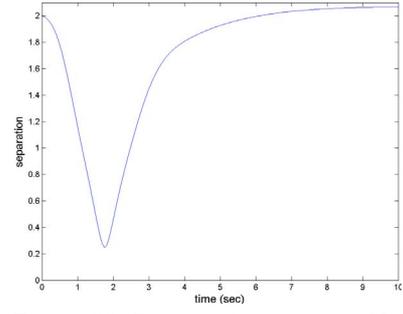

Figure-22: Distance between the UAVs

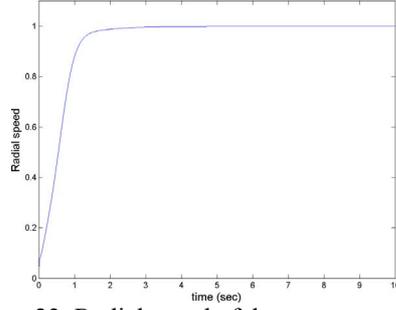

Figure-23: Radial speed of the non-cooperative UAV.

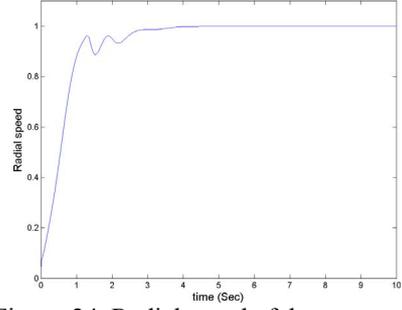

Figure-24: Radial speed of the maneuvering UAV.

*The redundant actuators case:*

The ability of the controller to deal with high redundancy in the actuation is demonstrated using the following simulation example. A spherical system is used with six control inputs (56,57). The parameters of the controller are: $K_\lambda=1$, $K_u=1$, $x(0)=y(0)=z(0)=0$, $v(0)=0$, $\theta(0)=\pi/2$, $\phi(0)=\pi/2$, $x_T=y_T=z_T=2$. The radial speed of the system is required to be as close as possible to $v_r=1$. Figure-25 shows the 3-D spatial trajectory and figure-26 shows the corresponding xyz components. As can be seen the trajectory converged to the target in a well-behaved manner. It can also be seen that the en route radial velocity (figure-27) converge to the desired radial velocity. The local orientation angles are shown in figure-28. The six control signals are shown in figure-29. As can be seen the signals are well-behaved.

$$\begin{aligned}
\dot{x} &= v \cdot C\varphi C\theta \\
\dot{y} &= v \cdot S\varphi S\theta \\
\dot{z} &= v \cdot C\theta \\
\dot{v} &= u_1 + u_4 \\
\dot{\theta} &= u_2 + u_3 + u_5 \\
\dot{\varphi} &= u_2 + u_4 + u_6
\end{aligned} \quad (56)$$

$$J_\lambda = \begin{bmatrix} C\varphi S\theta & -v \cdot S\varphi S\theta & v \cdot C\varphi C\theta \\ S\varphi S\theta & v \cdot C\varphi S\theta & v \cdot S\varphi C\theta \\ C\theta & \mathbf{0} & -v \cdot S\theta \end{bmatrix}$$

$$J_u = \begin{bmatrix} 1 & 0 & 0 & 1 & 0 & 0 \\ 0 & 1 & 1 & 0 & 1 & 0 \\ 0 & 1 & 0 & 1 & 0 & 1 \end{bmatrix} \quad (57)$$

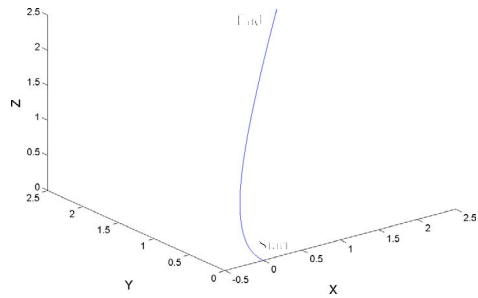
Figure-25: Spatial trajectory

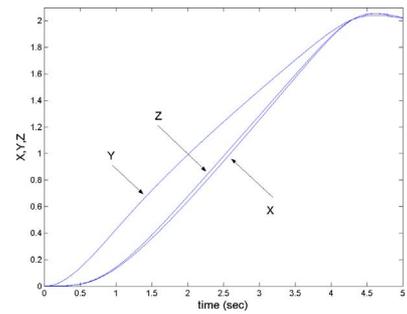
Figure-26: x,y,z trajectory component

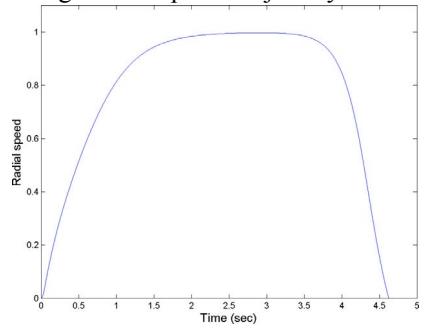
Figure-27: Radial speed

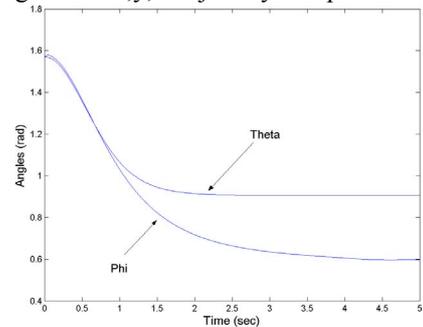
Figure-28: Orientation angles

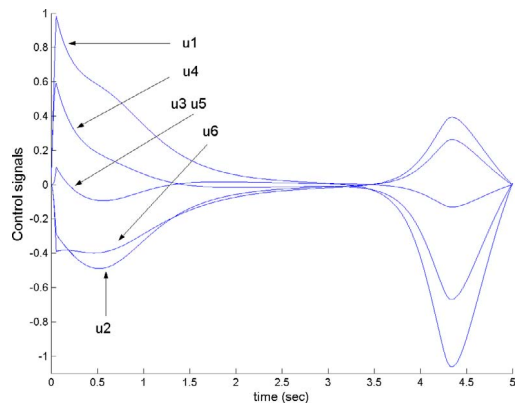
Figure-29: Control inputs.

The same example in figure-25 is repeated with the absolute values of the control components constrained not to exceed 0.4. As can be seen from figures-30,31, the trajectory still converged to the target point in a well-behaved manner. The control components are shown in figure-32; they are well-behaved and strictly restricted to the desired region.

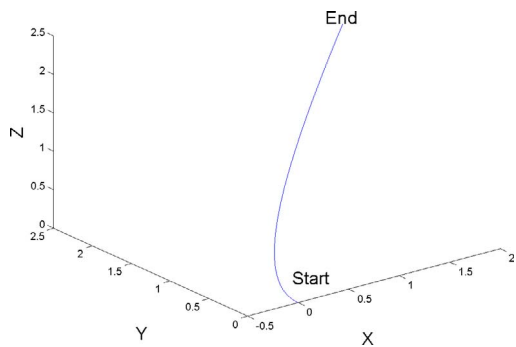
Figure-30: 3D trajectory - constrained.

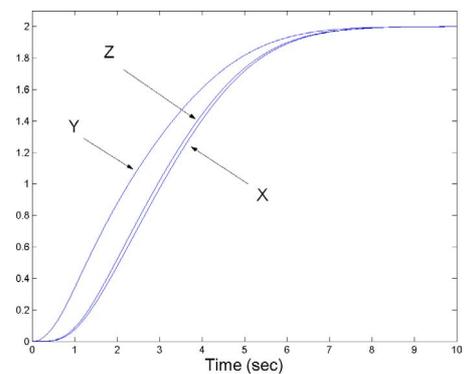
Figure-31: X,Y,Z trajectory components - constrained.

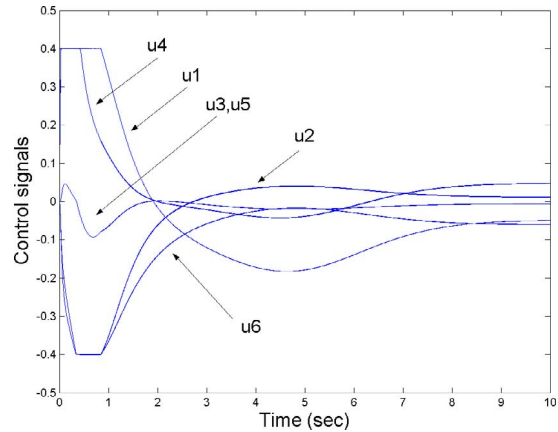

Figure-32: Constrained control signals.

While for this case the constrained system converged to the same target point as the unconstrained system, the constrained control signal converged to a different value than the unconstrained signal. This is caused by the redundancy in the actuation and the fact that an infinite number of solutions for the actuation part in 56 could lead to zero motion of the local state at steady state (figure-33).

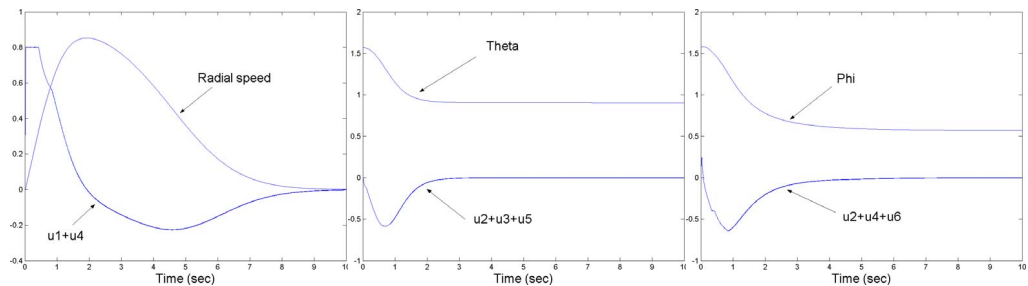

Figure-33: Vanishing steady state actuation control - constrained system.

Constraining the controller does not limit the ability of the spherical UAV to perform relatively involved maneuvers. In the following example the UAV described by equation (56) is required to climb up to a height of z=2 and perform a spiral maneuver in the x-y plane typically used in search patterns. The magnitude of each control signal is required not exceed 0.4. Also a unity radial speed is required. As can be seen from the 3D trajectory and its projection onto the x-y and y-z planes (figure-34), the controller was able to make the UAV efficiently perform the maneuver while keeping the control signals (figure-35) bounded and well-behaved. The radial speed is well-behaved and settles on the desired value (figure-36). The local orientation angles ($\phi$ and $\theta$) are also well-behaved (figure-37).

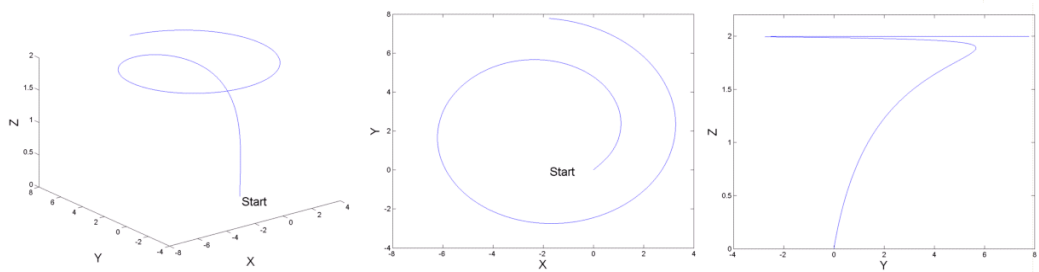

Figure-34: The UAV 3D trajectory and its projection onto the X-Y and Y-Z planes

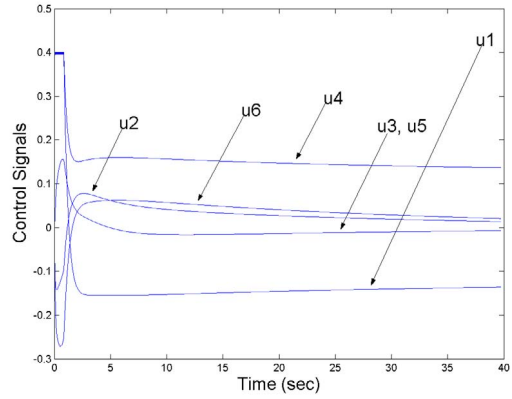

Figure-35: The constrained control signals

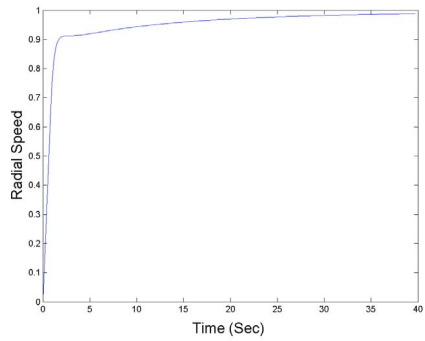
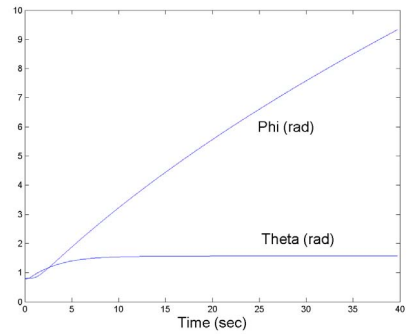

Figure-36: The radial speed of the UAVFigure-37: The orientation angles of the UAV

In the following example, the robustness of the control scheme is tested by adding noise to the control signals from the previous example. As can be seen, the trajectory remained well-behaved (figures-38,39). The noise effect on the radial velocity and the orientation is minimal (figures-40,41). The noisy control signal u1 is shown in figure-42.

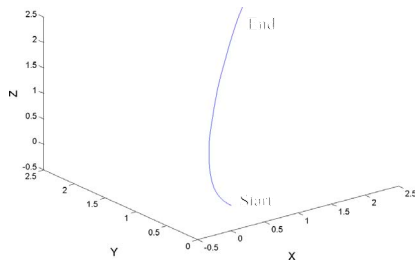
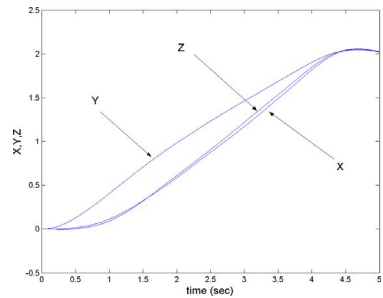

Figure-38: Noisy spatial trajectory,Figure-39: Noisy x,y,z components

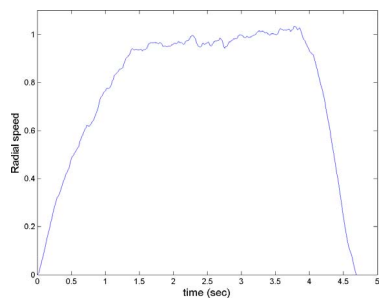
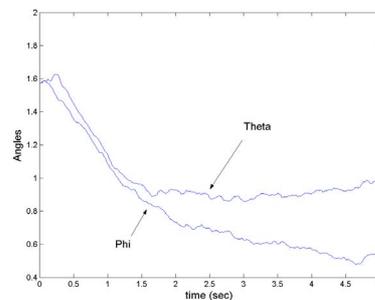

Figure-40: Noisy radial speedFigure-41: Noisy orientation angles

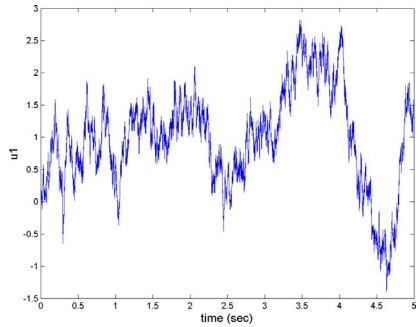
Figure-42; Noisy u1 control signal.

*Dynamic and kinematic trajectories compliance :*
This example demonstrates the ability of the suggested controller to make the dynamic trajectory comply with the kinematic one generated from a harmonic potential field planner. According to proposition-3, if the initial conditions are correctly set, it is possible to make the two trajectories identical hence enabling the dynamic system to enforce all the constraints encoded in the kinematic path. The G-Harmonic potential method used to plan trajectories in ambiguous non-divisible environments [38,39] is used to lay a kinematic path between two points in the X-Y projection of the UAV space. The X-Y environment is described by an intensity map where high intensity implies good fitness for motion to pass trough while low intensity (dark regions) implies low fitness. The suggested controller is required to fly the redundant, spherical system (56) in 3D space at an altitude Z=2 using the XY information from the G-harmonic potential to move the UAV in the X-Y plane from the start point to the target. The magnitude of the control components is required not to exceed 0.4 and the reference radial velocity is 1. The initial values of the local UAV variables are: $v(0)=0$, $\theta(0)=\phi(0)=\pi/4$. The controller was able to successfully drive the UAV to its target. Figure-43 shows the 3D trajectory and figure-44 shows the trajectory projection on the X-Z plane.

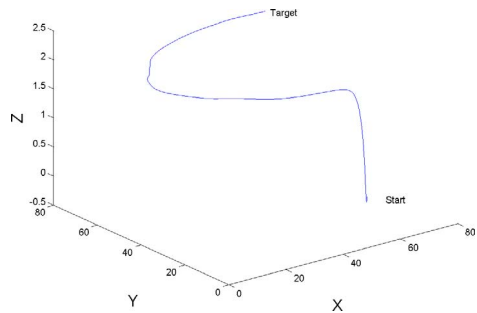
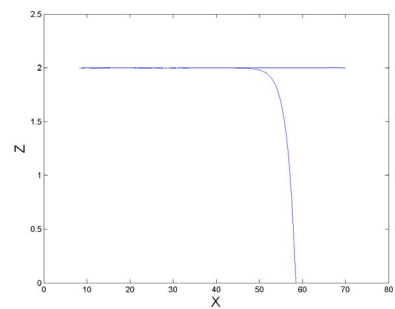

Figure-43: 3D trajectory  Figure-44: X-Z trajectory projection

Figure-45 shows the projection of the trajectory in the X-Y plane superimposed on the environment. The solid line represents the dynamic trajectory and the dotted line represents the kinematic trajectory. As can be seen there is discrepancy between the two. In figure-46, the initial values of the local UAV variables are changed to $v(0)=0$, $\theta(0)=\pi/4$, $=\phi(0)=-.225\pi$ to make the initial start of the two paths as much close as possible. As can be seen the kinematic and dynamic trajectories become almost identical.

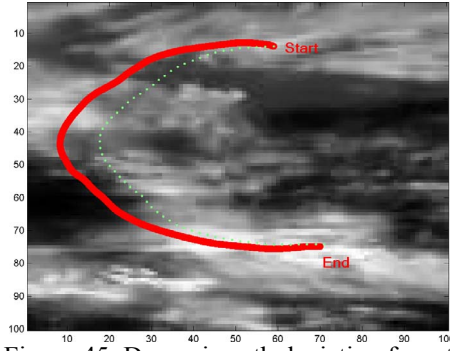 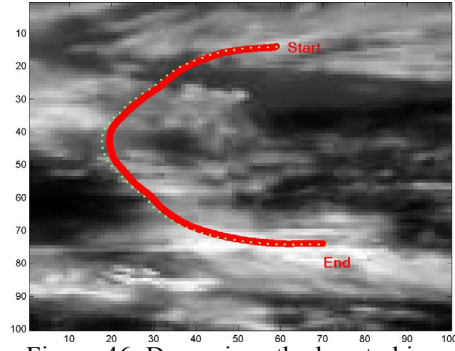

| Figure-45: Dynamic path deviating from the kinematic one | Figure-46: Dynamic path close to kinematic path |

*The underactuated case:*
The following example examines the behavior of the controller for the underactuated system described by equations (58,59). The number of control variables in the previous systems are reduced to only two which is less than the minimum needed to fully control the system in 3D space. The initial conditions, target and settings are the same as in the example shown in figure-25. As can be seen from figures 47,48 the system remained stable; however, it got trapped in a local minima failing to converge to the target.

$$\begin{aligned}\dot{x} &= v \cdot C\varphi C\theta \\ \dot{y} &= v \cdot S\varphi S\theta \\ \dot{z} &= v \cdot C\theta \\ \dot{v} &= u_1 \\ \dot{\theta} &= u_2 \\ \dot{\varphi} &= u_2 \end{aligned} \quad (58)$$

$$J_\lambda = \begin{bmatrix} C\varphi S\theta & -v \cdot S\varphi S\theta & v \cdot C\varphi C\theta \\ S\varphi S\theta & v \cdot C\varphi S\theta & v \cdot S\varphi C\theta \\ C\theta & 0 & -v \cdot S\theta \end{bmatrix}$$

$$J_u = \begin{bmatrix} 1 & 0 \\ 0 & 1 \\ 0 & 1 \end{bmatrix} \quad (59)$$

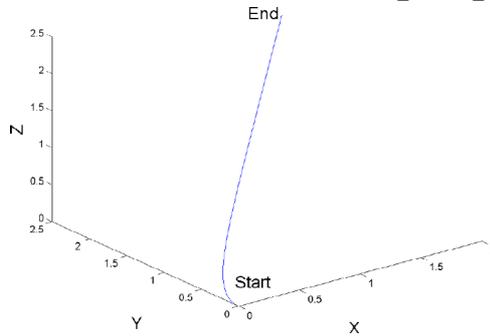 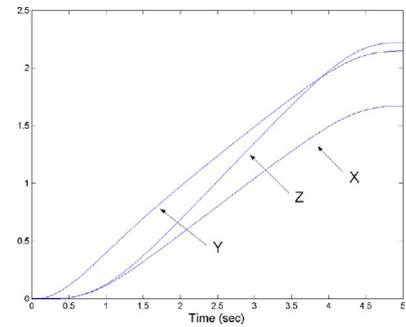

| Figure-47: Trajectory, underactuated UAV. | Figure-48: XYZ components, underactuated UAV. |

## VII. Conclusions

This paper demonstrates the ability of the harmonic potential field motion planning approach to deal with realistic planning problems such as the kinodynamic planning of motion for a UAV. Although the suggested solution is relatively simple (compared to the existing approaches) it amasses several important features desired for planning motion for a UAV. The structure of the controller is simple, making it highly possible to implement using inexpensive hardware. Despite this simplicity, the controller can tackle the exact model of a wide class of UAVs and provide an unconditionally stable, easy to tune response. The ability to effectively migrate, in a provably-correct manner, the kinematic path characteristics from an HPF-based planner to the dynamic trajectory of a UAV makes it possible to impose a wide class of constraints that accommodate the demands that a realistic environment imposes. It is worth noting that the manner in which the controller is constructed has the potential to tackle the full nonlinear model of a UAV with the rudder and aileron systems included [48,pp.103]. The work in this paper clearly demonstrates the promising potential the HPF approach has and its applicability to real situations.

*Acknowledgment:* The author acknowledges the assistance of King Fahad University of Petroleum and Minerals in supporting this work.